\documentclass{article}
\usepackage{iclr2027_conference,times}

\usepackage{amsmath,amsfonts,bm}

\def\eqref#1{equation~\ref{#1}}

\def\1{\bm{1}}

\DeclareMathAlphabet{\mathsfit}{\encodingdefault}{\sfdefault}{m}{sl}
\SetMathAlphabet{\mathsfit}{bold}{\encodingdefault}{\sfdefault}{bx}{n}

\usepackage{url}

\usepackage[T1]{fontenc}
\usepackage[utf8]{inputenc}
\usepackage{lmodern}
\usepackage[top=25mm,bottom=25mm,left=25mm,right=25mm]{geometry}
\usepackage{amsmath,amssymb,amsthm,mathtools}
\usepackage{booktabs}
\usepackage{array}
\usepackage{tabularx}
\usepackage{microtype}
\usepackage{enumitem}
\usepackage{textcomp}
\usepackage{tcolorbox}
\tcbuselibrary{most}
\tcbuselibrary{skins,breakable}
\usepackage{needspace}

\theoremstyle{definition}

\theoremstyle{plain}

\usepackage[ruled,vlined]{algorithm2e}
\usepackage{amsfonts}
\usepackage{amsmath}
\usepackage{amssymb}
\usepackage{amsthm}
\usepackage{bm}
\usepackage{booktabs}
\usepackage[T1]{fontenc}
\usepackage{graphicx}
\usepackage{inconsolata}
\usepackage[utf8]{inputenc}
\usepackage{latexsym}
\usepackage{makecell}
\usepackage{microtype}
\usepackage{multirow}
\usepackage{subcaption}
\usepackage{wrapfig}
\tcbset{
  promptbox/.style={
    enhanced,
    breakable,
    colback=white,
    colframe=black!40,
    boxrule=0.5pt,
    arc=2pt,
    left=6pt,right=6pt,top=6pt,bottom=6pt,
    fonttitle=\bfseries,
    title={#1},
  }
}
\usepackage{times}

\usepackage{pifont}
\newcommand{\cmark}{\ding{51}}
\newcommand{\xmark}{\ding{55}}

\usepackage{adjustbox}
\usepackage{hyperref}

\setcitestyle{authoryear,round,semicolon,aysep={,}}

\title{PoliticalSim: Multi-Layer Political Dynamics Simulation by LLM Agents}

\author{Eri Onami  \\
Institute of Science Tokyo\\
\texttt{eri.onami at nlp.comp.isct.ac.jp} \\
\And
Takehiro Takayanagi \\
The University of Tokyo, Simulacra Inc.\\
\texttt{takayanagi-takehiro590 at g.ecc.u-tokyo.ac.jp} \\
\And
Ryutaro Mori \\
RIKEN\\
\texttt{ryutau.mori at gmail.com} \\
\And
Ryuji Hashimoto \\
The University of Tokyo, Simulacra Inc.\\
\texttt{hashimoto-ryuji419 at g.ecc.u-tokyo.ac.jp} \\
\And
Kyung-Gu Lee\\
Independent Researcher\\
\texttt{byzantion at snu.ac.kr} \\
\And
Naoaki Okazaki \\
Institute of Science Tokyo\\
\texttt{okazaki at comp.isct.ac.jp} \\
}

\iclrfinalcopy
\begin{document}

\maketitle
\lhead{Preprint}

\begin{abstract}
Traditional political science analyzed political decision making processes, including voters’ electoral choices, legislative deliberation by selected representatives in parliament, and institutional interaction between parliament and other actors such as the bureaucracy and interest group, by modeling economically rational individuals who act to maximize their utility.
These political theories are grounded in mathematically rigorous economic theories such as game theory, neglecting individual's diversity and social interactions, lacking controlled experiments because randomized intervention in real societies is almost impossible.
Advances in artificial intelligence have made it possible to reproduce patterns of human interaction that mathematical theory alone could not capture by using Agent-Based Modeling.
However, it remains unclear to what extent LLMs can comprehend complex political institutions and replicate human political behavior, and prior work didn't provide a simulation environment for controlled experiments that enable or disable individual mechanisms while holding other conditions constant.
In this study, we propose PoliticalSim, a novel framework for simulating the political activities of voters, parliament representatives, and institutional interactions, enabling controlled experiments that enable and disable individual political factors while holding other conditions constant.
By this framework, we show that LLMs can reproduce some important phenomena observed in political science in both hypothetical and real world grounded settings, and we experimentally identify which factors affect the final outcomes.

\end{abstract}

\section{Introduction}

In political science, democratic political system is analyzed by utilizing mathematically grounded theories of economics to examine the behavior of the actors involved in elections, such as voters and politicians.
Assuming that voters and politicians are rational individuals who seek to maximize their own utility, theoretical approaches of political science employed the application of rational choice theory to analyze actor's decision making~\citep{Black1948,Downs1957AnET,Riker_Ordeshook_1968}.
It has been pointed out, however, that human behavior is not solely rational.
Numerous political phenomena are considered to be the caused by, denominational religious behavior, the urban and rural divide, the division between capital and labor~\citep{LipsetRokkan1967}, ethnicity~\citep{Horowitz1985}, party identification~\citep{Campbell_1960}, group identity~\citep{achen2016democracy} and social norms and conformity pressure~\citep{GERBER_GREEN_LARIMER_2008}, that cannot be explained by economic theory alone.
It is impossible to determine which factor constitutes the dominant cause because the outcomes of elections and legislative deliberations are always formed by multiple interacting factors.
Moreover, since political circumstances and social problems are constantly evolving and it is difficult to compare outcomes under identical conditions, where analyzing causal relationships through controlled comparison, holding all conditions constant except a single variable.
These two factors represents the greatest challenge in political science research.
Large language model (LLM) based simulations, however, may enable controlled variation in institutions and conditions to examine whether the behaviours and outcomes predicted by political science theories emerge.

In recent years, social simulations based on LLM agents have been proposed.
These approaches utilize the capacity of LLMs to comprehend long contexts, reason logically, emulate human-like behavior, and interact with other agents through natural language~\citep{Aher2023a, xie2024canllm, Piatti2024, weng2025benchform, zhu-etal-2025-conformity}.

In political science domain, several approaches to leverage natural language processing and LLMs are proposed, such as extracting distributions of voter opinion from social media text to analyze elections~\citep{yano-etal-2012-textual}, analyzing legislative deliberation in state and national legislatures from the text of submitted bills~\citep{zhang2026policon}, simulating legislative deliberation using LLMs~\citep{Lietal2025_politicalactoragent}.
Prior studies have used social media posts, the texts of bills debated in legislatures, or Wikipedia article of legislators to enable simulated agents to mimic human behaviour.
However, prior works didn't examine whether LLM can understand complex political mechanisms, act as human political behavior and interact with other agents, and reproduce real-world observations.

To address this challenge, we simulate political phenomena in three stage simulations of electoral voting, parliamentary deliberation, and institutional interaction, to examine whether LLM agents can simulate political activities and produce outcomes consistent with real-world political patterns.
The proposed framework simulates both hypothetical political processes and those grounded in real-world settings across seven countries, the United States, the United Kingdom, Germany, France, Italy, South Korea, and Japan.
By enabling or disabling mechanisms derived from political science theories while holding other conditions constant, it allows controlled investigation of causal relationships within the simulation.

Our contributions are as follows:
\begin{itemize}
    \item We propose PoliticalSim, a new framework for simulating political processes, including elections, legislative deliberation, and interactions between legislatures, bureaucracies, and interest groups. PoliticalSim does not require additional model training or detailed agent profiles derived from social media posts or Wikipedia to ground the simulations in the real world.

    \item PoliticalSim incorporates mechanisms drawn from political science theories and allows individual mechanisms to be enabled or disabled while holding other conditions constant, enabling comparisons of how well competing theories explain simulated political outcomes.

    \item We evaluate LLM agent behaviour against established political science theories through a three-stage simulation of political dynamics, examining whether LLM agents can reproduce diverse phenomena documented in empirical political science.
\end{itemize}

\section{Related Work}

\textbf{LLM based Social Simulations.} Recent studies suggest that LLMs offer a foundation for large-scale social simulation by supporting psychologically grounded behavior, heterogeneous personas, and natural language based interactions.
Early work on Homo silicus explored the use of LLMs as simulated participants in economic experiments, examining whether their decisions reproduce patterns observed in human behavior ~\citet{10.1145/3670865.3673513}.
Moving from individual decisions to macroeconomic dynamics, EconAgent~\citep{li-etal-2024-econagent} models heterogeneous agents whose labor and consumption decisions reflect personal experiences and changing market conditions.
\citet{3709347.3743903} extend this approach with an LLM-based policy planner that adjusts tax rates and inflation targets in response to economic conditions and shocks.
Other studies investigate how interactions among economic agents generate collective outcomes.
TwinMarket simulates financial markets to examine how individual decisions and social feedback produce market dynamics~\citep{yang2025twinmarket}, while \citet{10.1145/3711896.3736929} model online game economies, demonstrating emergent specialization and price fluctuations.

\textbf{Simulation in Political Science.}
Several dataset benchmarks are presented to simulate parliamentary deliberation such as US local parliament~\citep{davoodi-goldwasser-2024-analysis} EU parliament~\citep{zhang2026policon}, and Greek parliament~\citep{Dritsa2022} legislative process by real legislative text inputs.
Political Actor Agent~\citep{Lietal2025_politicalactoragent} proposed a framework that models real-world legislators as LLM agents to predict their votes in favor of, against, or abstaining on bills.
The study examined detailed legislator profiles from Wikipedia, reasoning that legislators are both representatives of the district and supporters of a political party, and the influence of other leading legislators.
However, it did not simulate the full legislative process, including committee deliberations, interactions with bureaucrats, the drafting of amendments, and negotiations with interest groups, thereby oversimplifying real-world political procedures.

\section{Political Simulation}

\begin{figure*}[tbp]
  \centering
  \includegraphics[width=0.95\linewidth]{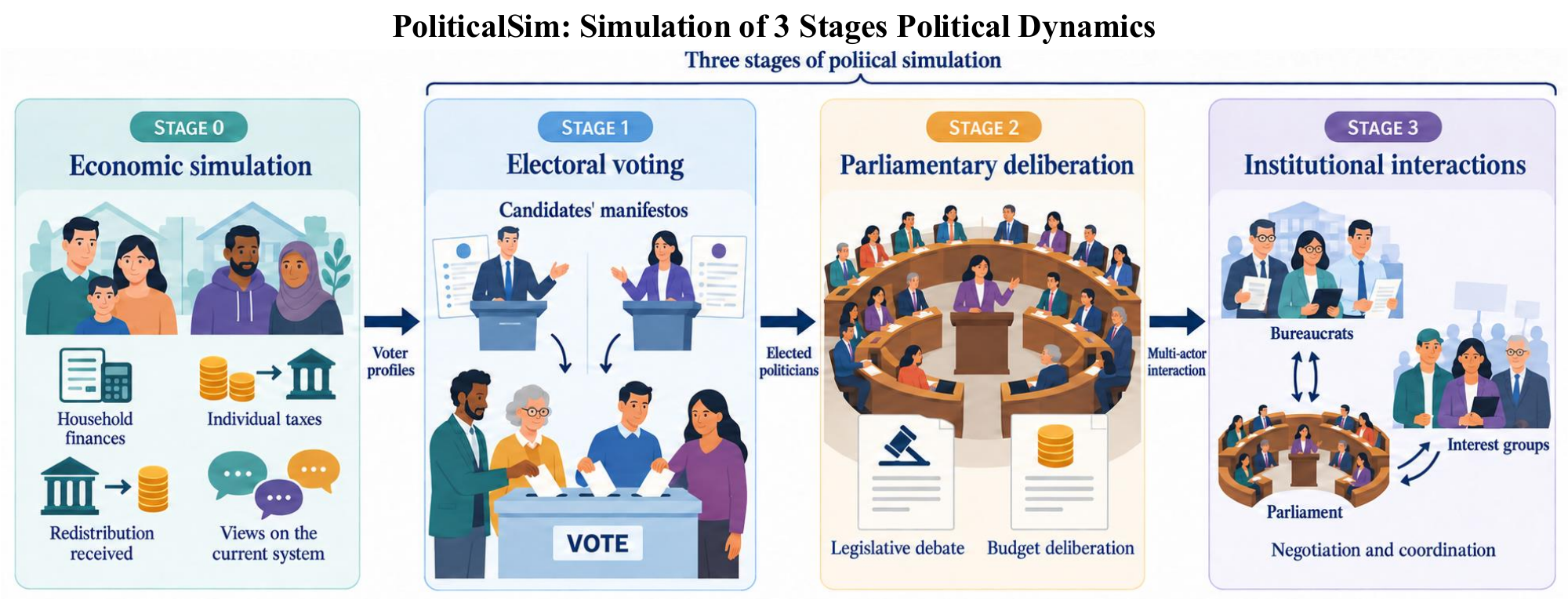}
  \caption{Overview of three stages political simulation, namely, Electoral voting, Parliamentary deliberation, and  Institutional interactions, based on economic simulation.}
  \label{Fig:teaser}
\end{figure*}

As shown in Figure~\ref{Fig:teaser}, namely, we assume that political phenomena consist of three stages: (i) voting behavior through which voters elect politicians, (ii) activities conducted within the legislature by elected politicians, including budgetary and legislative deliberation, the selection of the prime minister, and committee activities, and (iii) interactions between the parliament and institutive actors such as bereaucrats and interesting groups.

\subsection{Stage 0: LLM Agent-Based Economic State Simulation of voters}
\label{sec:economy}
A substantial body of prior political science research has shown that voters’ income levels, macroeconomic conditions, and tax and redistributions are strongly associated with individual voting behavior\citep{Meltzer_Richard_1981,ROMER1975,Kramer1971, Fiorina1981RetrospectiveVI, Kinder_Kiewiet_1981,GelmanBafumiPark2007,ALESINA201193}.
Therefore, as a prerequisite for simulating political dynamics, we simulate the economic conditions of society, including income, taxation, and redistribution.
We use ~\citet{hashimoto-etal-2026-heard} to provide a simulation that models individual income, taxation, and redistributive transfers and aggregates the distribution of opinions associated with agents’ economic circumstances.

\subsection{Stage 1: Electoral Voting}
\paragraph{Candidate policy representation.}
We construct candidate platforms at distinct positions along a common conservative and progressive axis. The hypothetical setting compares single member plurality, single non-transferable vote (SNTV) with district magnitude $M\in\{2,3,4\}$, and five seat proportional representation (PR) using D'Hondt allocation. Under SNTV, each voter casts one vote and the top $M$ candidates win. PR provides a comparison with different electoral incentives; it does not eliminate strategic voting.

\paragraph{Electoral evaluation.}
We evaluate the result in six aspects: consistency with the $M+1$ rule, the effective number of candidates (ENC), the effective number of parties (ENP), vote--seat disproportionality, agreement in party rankings, and identification of the party winning the most seats. With candidate vote shares $f_j^{(t)}$ and party vote shares $q_p^{(t)}$ expressed as fractions, we compute~\citep{LaaksoTaagepera1979}
\begin{equation}
\mathrm{ENC}^{(t)}=\left[\sum_j(f_j^{(t)})^2\right]^{-1},\qquad
\mathrm{ENP}^{(t)}=\left[\sum_p(q_p^{(t)})^2\right]^{-1}.
\label{eq:enc}
\end{equation}
Here, $t=1,2$ denotes voting before and after receiving poll information. The $M+1$ rule~\citep{Duverger1954,cox1997making} predicts that strategic coordination limits the number of viable candidates to at most $M+1$ under the relevant electoral conditions. We use ENC to describe vote concentration.

We measure vote and seat disproportionality using the Gallagher index,
$G=\sqrt{\frac12\sum_{p=1}^{N_{\mathrm{party}}}(V_p-S_p)^2}$,
where $V_p$ and $S_p$ are party $p$'s vote and seat shares in percentages.
Zero indicates perfect proportionality and larger values indicate greater discrepancies.
We assess party rankings using Spearman's $\rho$ between simulated and observed seat shares, assigning average ranks to ties.
Values of $1$, $-1$, and $0$ indicate identical rankings, reversed rankings, and no rank correlation, respectively.
For the United States, where only two parties are included, we report whether the party winning the most seats is correctly identified, since $\rho$ provides no additional ranking information in the absence of ties.

\subsection{Stage2: Parliamentary Deliberation}
We run the agenda contains 100 fictional bills for seven countries, five in each of 20 policy fields, including social welfare, agriculture, and manufacturing. Each bill describes the problem, proposed measures, fiscal cost, and who gains or loses. Bill $b$ has an industry effect $\delta_{bh}\in[-2,2]$ for industry $h$. Its effect on district $d$ is $e_{db}=\sum_{h=1}^{H}w_{dh}\delta_{bh}$, where $H$ is the number of industries and $w_{dh}$ represents the district's industrial composition. Positive and negative effects indicate gains and losses, respectively.
Party $p$'s position, $\ell_{pb}\in\{-1,+1\}$, denotes opposition or support. It is generated once from the party platform and held fixed.

For bill a proposal $x$ with scale $s_x$, legislator $i$ from party $P(i)$ has payoff
\begin{equation}
u_i(x)=\begin{cases}
0.7s_x e_{d(i)b}+0.3s_x\ell_{P(i)b}-\kappa s_x^2,
&\text{district-elected legislators},\\
s_x\ell_{P(i)b}-\kappa s_x^2,
&\text{list-elected legislators},
\end{cases}\qquad \kappa=0.5.
\end{equation}
Here, $d(i)$ is legislator $i$'s constituency.
For district elected legislators, the model assigns weights of 0.7 and 0.3 to constituency interests and the party position, respectively, reflecting the importance of constituency support for reelection.
These weights are modeling assumptions.
List-elected legislators have no constituency interest term.

\subsection{Stage 3: Institutional Interaction}
\paragraph{The Interest Groups.}
The chair's policy valuation is
$g(x)=\lambda\sum_{i\in\mathcal C}u_i(x)$,
where $\mathcal C$ contains all nine committee members.
The chair is included with the same weight as the other members.
Let $\mathcal L$ denote the interest groups and $D_\ell(x)$ the contribution offered by group $\ell$ conditional on selecting $x$. For the selected proposal $\hat x$, the chair and group payoffs are $U_{\mathrm{chair}}=g(\hat x)+\sum_{\ell\in\mathcal L}D_\ell(\hat x)$ and $U_\ell=v_\ell(\hat x)-D_\ell(\hat x)$, respectively.
Here, $v_\ell(x)$ denotes interest group $\ell$'s valuation of proposal $x$ in monetary units, before subtracting its contribution. We set $v_\ell(x)=\nu_\ell s_x$, where $s_x$ is the proposal's scale and $\nu_\ell$ is the group's gain or loss per unit scale: $\nu_\ell>0$ for the benefiting group and $\nu_\ell<0$ for the adversely affected group.
We define the theoretically efficient proposal $x^*$ as a proposal that maximizes the sum of the committee's and interest groups' valuations:
\begin{equation}
x^*\in\arg\max_x\left[g(x)+\sum_{\ell\in\mathcal L}v_\ell(x)\right].
\end{equation}

Interest group valuations are proportional to proposal scale:
$v_+(x)=V_+s_x$ for the benefiting group and
$v_-(x)=-V_-s_x$ for the adversely affected group.
The magnitudes $V_+$ and $V_-$ are determined separately from
committee members' exposure to positive and negative industry effects
and need not be equal.
A group is included only if the bill benefits or harms at least one
industry, respectively.

\paragraph{The Bureaucracy.}
A bureau and a parliamentary committee negotiate programme quantity $Q\geq0$ and budget $T$.
Bills reported at scale $s>0$ with medium or high fiscal cost enter negotiations.
The bureau knows benefits and costs but the committee receives benefits and an unverified cost report.
The committee's benefit and implementation cost are
\begin{equation}
H_C(Q)=a_CQ-\eta Q^2,\qquad C(Q)=cQ+\zeta Q^2.
\end{equation}
Here, $a_C$ and $c$ are initial marginal benefit and cost, and $\eta,\zeta\geq0$ govern declining marginal benefit and increasing marginal cost.
This follows the linear marginal-benefit/cost structure of Niskanen~\citep{niskanen1968}.
For $a_C>c$ and $\eta+\zeta>0$, the quantity maximizing $H_C(Q)-C(Q)$ is $Q_C^{\mathrm{opt}}=(a_C-c)/[2(\eta+\zeta)]$.
This benchmark uses the committee's valuation, not aggregate social welfare.
The committee's surplus is $H_C(Q)-T$ and the bureau's discretionary budget (slack) is $T-C(Q)$.
To model the bureau's preferences for providing more public services
and retaining a budget beyond implementation costs~\citep{migue1974}, we use the theoretical objective $U_{\mathrm{bur}}(Q,T)=Q^\beta[T-C(Q)]^\gamma$ on feasible allocations $T\geq C(Q)$, with $\beta,\gamma\geq0$ and $\beta+\gamma>0$.
In this theoretical benchmark, the bureau makes a single offer that the committee can only accept or reject.
The committee accepts if its benefit $H_C(Q)$ is at least as large as the requested budget $T$, including when they are equal.
The bureau therefore requests the maximum acceptable budget, $T=H_C(Q)$. Maximizing its utility under this condition gives
\begin{equation}
\frac{Q_{\mathrm{bur}}}{Q_C^{\mathrm{opt}}}
=\frac{2(\beta+\gamma)}{\beta+2\gamma}\in[1,2].
\end{equation}
When $\beta=\gamma\geq0$, $Q_{\mathrm{bur}}/Q_C^{\mathrm{opt}}$ is $4/3$.

\section{Experimental Settings}
We use latest open-weight models with reasoning abilities and knowledge for applied domain including mathmatics, economics, and politics: Llama-3.1-8B-Instruct~\citep{grattafiori2024llama3herdmodels}, Qwen3-14B~\citep{qwen3technicalreport}, Gemma-4-12B-it~\citep{gemmateam2026gemma4technicalreport}, and gpt-oss-20b~\citep{openai2025gptoss120bgptoss20bmodel}.

\subsection{Hypothetical and Real World Election Simulations}

\paragraph{Hypothetical elections.}
For each LLM, we conduct ten runs with 100 household agents per run. Voters receive personas, economic circumstances, cumulative taxes and transfers, and socioeconomic opinions from the economic simulation. Seven hypothetical parties represent positions $\alpha_j\in[0,1.75]$ on a common one-dimensional policy scale. Each electorate votes under five systems, single member plurality ($M=1$), single non-transferable vote (SNTV; $M=2,3,4$), and proportional representation (PR), with five seats allocated by D'Hondt.
Alternatives are presented in random order for each voter and round.
Voters first vote without observing others' choices, providing a preference ranking and explanation.
They then vote again after receiving the first round totals and a reminder of their previous choice.
Under plurality and SNTV, the prompt discusses wasted votes, permits strategic voting, and identifies the top $M$ candidates as likely winners.
Under PR, it allows consideration of the poll without identifying likely winners.
The comparison therefore captures responses to both polling information and system-specific guidance.

\paragraph{Real World elections.}
We simulate lower house elections in the United States, the United Kingdom, France, Germany, Italy, South Korea, and Japan using the configured election years and electoral rules.
Economic histories are fixed across voting models, using 800 household agents from eight completed economic simulation runs.
Let $U$ denote the number of simulated electoral districts in a country.
Groups of 100 agents are assigned cyclically to electoral units, generating $100U$ initial voting decisions for a country with $U$ units.
For each district unit, we input an ideological leaning score from $\mathcal N(0,0.8^2)$, and convert it into a verbal description ranging from much more progressive to much more conservative than the national average.
Religious affiliations are assigned to voters using demographic data on each country's religious composition by using a seeded random permutation and included in their prompts.
The parties are extracted from actual data on the previous election in each country.
Each selected party is classified as progressive, centrist, or conservative based on its political ideology.
Each party’s platform comprises two real political issues from its country, with its position for or against each issue determined by its assigned political ideology.
Party names are replaced with anonymous labels in voter agent input.
Voters receive previous election national vote shares and district seat counts, together with profiles describing their locality, policy concerns, religion and habitual party support.
Each electoral district is assigned information on its urban or rural character and local industries.
Habitual support is rescaled towards prior national vote shares.
Religion and habitual support are assigned by the largest remainder method with seeded permutations.
Country specific modules implement plurality in the United States and United Kingdom, parallel systems in Japan and Italy, and compensatory or partially compensatory systems in Germany and South Korea, with configured proportional allocation methods and thresholds.
France uses a simplified two round procedure.
If no party obtain majority in first round, the leading two parties and additional parties with at least 12.5\% of simulated votes remain.
We collected 10,000 voter's result from each seven country across seven models.

\subsection{Parliament Delibaration Simulation with the Bureaucracy and the Interesting Group Interaction}

Bills proceed through committee deliberation and proposal selection, interest-group briefings, budget negotiations, contribution offers to individual legislators, and plenary deliberation and voting.
Bills that terminated in committee ($s=0$) do not proceed to later stages.

\paragraph{Committee.}
District-elected legislators are instructed to consider constituency interests and party positions, while list-elected legislators are instructed to consider party positions.
Legislators are given information on the value of each policy area $a_{ik}$, its effects on their constituency $e^{\mathrm{field}}_{ik}$, and their party’s priorities $\omega_{P(i)k}$.
Each legislator specifies their first and second choices of policy areas in which they wish to deliberate on bills, and committees are formed according to these preferences.
During committee deliberation, each member first communicates to the other members their preferred bill scale, to maximize their own interests.
The bill scale $s$ denotes termination in committee at $s=0$, a reduction in scope at $s=0.25$–$0.75$, the original proposal at $s=1$, and an expansion in scope at $s\geq1.25$.
Based on members' views and the chair's objective, the chair selects $s\in\{0,0.25,0.50,0.75,1.00,1.25,1.50\}$. A choice of zero terminates the bill in committee and a positive scale reports it to the plenary session in parliament.

\paragraph{Budget negotiations with the bureaucracy.}
For bills reported by the committee with medium or high fiscal cost, the committee and implementing bureau negotiate quantity $Q$ and budget $T$ over at most five rounds.
In round $r$, the bureau director accepts the chair's previous counteroffer or proposes $(Q_r,T_r,\widehat C_r)$, namely, quantity, requested budget, and reported cost.
In the first round only, the nine committee members see their individual valuations $a_{ik}Q_r-\eta Q_r^2$ and constituency budget burdens $\tau_iT_r$, where $\tau_i$ is the budget share borne by legislator $i$'s constituency.
Legislators submit preferred quantities, maximum acceptable budgets, and brief justifications.
Given the bureau's proposal and member's report, the chair approves, counters with $(Q'_r,T'_r)$, or rejects.
Approval ends negotiations or a counteroffer can be accepted in the next round.
A counteroffer in the fifth round cannot be accepted.
The bureau is notified when $T_r<C(Q_r)$.
If no agreement is reached within five rounds, the policy is not implemented and $(Q,T)=(0,0)$.

\paragraph{Interest group briefings and contribution offers.}
After the committee reports a bill, groups representing benefiting and adversely affected industries provide information to selected legislators without offering money.
Each group selects up to 12 recipients since limiting the number of recipients allows us to observe whether interest groups target allies, neutral legislators, or opponents~\citep{HALL_DEARDORFF_2006}.
For recipient selection, interest group can observe the candidates' true constituency effects at the reported scale and their party positions.
Interest groups then provides each selected recipient with a claimed effect $\widehat c_{\ell ib}\in[-2,2]$ and a short message.
Interest groups may provide inaccurate information that favours their own position rather than truthfully reporting the bill’s effects.
The benefiting group is processed first, but neither group sees the other's briefings.
A legislator may receive messages from both groups.
Briefings are included in the recipients' plenary session inputs and, for committee members.

After budget negotiations, the same industry groups may offer contributions to individual legislators' campaign funds.
The benefiting group first offers amounts $A_i^+\geq0$ in exchange for support.
The adversely affected group observes these offers and then offers amounts $A_i^-\geq0$ in exchange for opposition.
Each interest group has a fixed contribution budget, limiting how much it can offer and to whom.
Legislators see the offers before plenary deliberation and voting. Under the payment rule used in the main experiment, a contribution is recorded only when the recipient casts the requested vote, irrespective of whether the bill passes and abstention yields no payment.

\paragraph{Plenary deliberation and voting.}
All legislators participate in two stages. First, each receives their persona, the bill and committee recommendation, party positions, any interest group briefings, budget-negotiation outcomes, and individual contribution offers.
District elected members also receive a rough estimate of the constituency effect,
$\widehat e_{ib}=\min\{2,\max\{-2,e_{ib}(s_b)+\epsilon_{ib}\}\}$,
where $\epsilon_{ib}\sim\mathcal N(0,0.5^2)$.
Each member states an initial position of support, oppose or undecided, and provides a brief explanation.
Second, each member receives their own initial statement, eight randomly sampled statements from other legislators, four from their own party and four from other parties, and the initial position for each party.
The member then votes to support, oppose or abstain and explains the reason of their decision.
A bill passes if support votes exceed opposition votes.
Abstentions are excluded from this comparison.
Enactment requires both reporting by the committee and passage on the plenary session.

\section{Results}
\label{sec:results}

\begin{table}[t]
\centering
\small
\caption{Average of Effective Number of Candidates before and after the opinion poll.}
\begin{tabular}{@{}lccccc@{}}
\toprule
 & \multicolumn{5}{c}{Effective Number of Candidates: round 1 $\to$ round 2} \\
\cmidrule(lr){2-6}
Condition  & SMD & $M{=}2$ & $M{=}3$ & $M{=}4$ & PR \\
\midrule
$M{+}1$ prediction & 2 & 3 & 4 & 5 & -- \\
\midrule
gpt-oss-20b  & 2.89$\to${2.45} & 2.91$\to${1.99} & 2.71$\to${2.30} & 2.85$\to${2.41} & 2.99$\to${2.56} \\
Llama-3.1-8B-Instruct  & 4.61$\to${1.48} & 4.61$\to${2.49} & 4.74$\to${1.96} & 3.93$\to${1.70} & 4.84$\to${2.44} \\
Qwen3-14B & 4.22$\to${2.27} & 4.22$\to${1.89} & 4.66$\to${2.89} & 3.90$\to${3.30} & 4.40$\to${2.20} \\
Gemma-4-12b-it & 2.59$\to${1.80} & 2.86$\to${1.94} & 2.90$\to${2.74} & 2.50$\to${2.68} & 2.95$\to${2.15} \\
\bottomrule
\end{tabular}
\begin{center}
\scriptsize
\setlength{\tabcolsep}{2.6pt}
\label{tab:enc}
\end{center}
\end{table}

\begin{table}[t]
\centering
\small
\caption{Result of real-world election simulation. $G_{\mathrm{real}}$ and $ENP_{\mathrm{real}}$ denote the observed Gallagher index and Effective Number of Parties, respectively. Values of $G$ and $ENP$ are shown in bold when their absolute differences from $G_{\mathrm{real}}$ and $ENP_{\mathrm{real}}$, respectively, are at most 1. Spearman’s $\rho$ values are also shown in bold when they are above 0.6.}
\label{tab:realworld}
\begin{tabular}{@{}lllrrrrcr@{}}
\toprule
Voting Model & Country & System & $G$ & $G_{real}$ & $ENP$ & $ENP_{real}$ & 1st Party & $\rho$ \\
\midrule
Gemma-4-12B-it & US & FPTP & \textbf{1.1} & 1.44 & 1.83 & 2.10 & \xmark & -- \\
 & UK & FPTP & 30.7 & 11.80 & 4.88 & 3.23 & \xmark & \textbf{0.69} \\
 & FR & two-round & 22.3 & 21.12 & 7.41 & 6.82 & \xmark & 0.07 \\
 & DE & MMP & 0.1 & 1.95 & 4.54 & 6.16 & \cmark & 0.39 \\
 & IT & parallel & 2.4 & 5.70 & 7.29 & 3.42 & \xmark & 0.23 \\
 & JP & parallel & 20.6 & 16.28 & \textbf{4.35} & 4.22 & \xmark & \textbf{0.71} \\
 & KR & semi-MMP & 19.7 & 13.65 & 3.07 & 3.10 & \xmark & \textbf{0.67} \\
\midrule
gpt-oss-20b & US & FPTP & 2.5 & 1.44 & \textbf{1.88} & 2.10 & \xmark & -- \\
 & UK & FPTP & 25.8 & 11.80 & \textbf{3.15} & 3.23 & \xmark & \textbf{0.69} \\
 & FR & two-round & 30.0 & 21.12 & 4.49 & 6.82 & \xmark & 0.51 \\
 & DE & MMP & \textbf{2.8} & 1.95 & 4.43 & 6.16 & \cmark & \textbf{0.70} \\
 & IT & parallel & \textbf{5.6} & 5.70 & 4.97 & 3.42 & \xmark & \textbf{0.66} \\
 & JP & parallel & 14.1 & 16.28 & \textbf{4.19 }& 4.22 & \cmark & \textbf{0.78} \\
 & KR & semi-MMP & 22.5 & 13.65 & \textbf{2.43} & 3.10 & \xmark & 0.41 \\
\bottomrule
\end{tabular}
\end{table}

\par
\Needspace{18\baselineskip}
\begin{wrapfigure}{r}{0.48\textwidth}
  \centering
  \includegraphics[width=\linewidth]{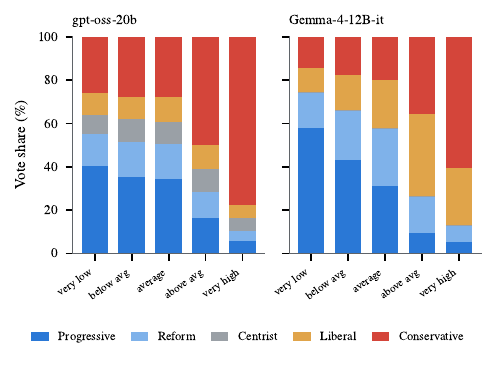}
  \caption{The result of voter agents chosen parties lie on the political spectrum from conservative to progressive. Voter agents are devided into five groups based on their economic circumstances.}
  \label{fig:figure_stage1_bracket_gptoss_gemma}
\end{wrapfigure}

\paragraph{Hypothetical and real world election.}
The Table~\ref{tab:enc} compares the voting without other voters' information and the voting that agents are notified information of other agents’ choices, their preferred candidate’s ranking among all candidates, and the likelihood of election given the district magnitude.
In single member districts, when agents are informed of other voters’ choices, the ENC ranges from 1.48 to 2.45 in all models.
Thus $M+1$ rule hypothesis in single-member district that the ENC $\approx2$ observed in the agent-based simulation.
Across models, Qwen3-14B produces results closest to the $M+1$ rule because ENC of round 1 doesn't shrink large in ENC of round 2 for M=1 to 4.
While the $M+1$ rule can be applied to proportional representation (PR) systems, PR with a large $M$ has a weak constraining power to incentivize voters from wasting votes, and the ENC is expected to remain close to that under sincere voting.
However, the models shrink ENC number in second round.
We divide voter agents into five groups based on their economic circumstances and visualise where their chosen parties lie on the political spectrum from conservative to progressive in Figure~\ref{fig:figure_stage1_bracket_gptoss_gemma}.

The Table~\ref{tab:realworld} shows the result of real world election system simulation across seven countries.
Even we didn't construct detailed agent profiles by matching them to real world social media posts or detailed Wikipedia entries of electoral candidates, gpt-oss-20b performs the best out of four models, with correctly identifying the party winning the most seats in two countries and achieving Spearman’s $\rho > 0.6$ for party rankings in four countries out of seven.
We only used statistical information, such as national industrial composition, constituency interests, and religion, together with result of the economic simulation, including income and tax distributions that approximate those of real economies.
By faithfully modelling each country’s political institutions, our framework may produce patterns and outcomes consistent with institutionalist theories, which emphasise how differences in institutions shape political behaviour and outcomes~\citep{Hall1996}.
Its ENP differs from the observed value by less than one point in four countries, while its Gallagher index falls within one point of the observed value in two.
Analyses of real world election data have shown that higher-income voters are more likely to support conservative parties, a tendency also reproduced by our models~\citep{Gelman2008}.

\paragraph{Simulation of Parliament Deliberation.}

\begin{table}[t]
\caption{Result of parliament deliberation in each process.
$s=0$ denotes the proportion of bills for which the committee chair’s final choice was rejection.
$s\neq0,1$ is the proportion of cases in which the final choice differed from the original proposal.
$=x^*$ is the proportion of cases in which the final choice matched the theoretically efficient proposal $x^*$, which maximizes the sum of the committee’s valuation, $a,g(x)$, and the two interest groups’ valuations, $\sum_i v_i(x)$.
$s-s^*$ is the mean difference between the scale $s$ of the final choice and the scale $s^*$ of the theoretically efficient proposal $x^*$.
Pass is the proportion of bills passed in the plenary session of the bills that reached the plenary session.
Enacted means the proportion of all bills that passed in the plenary.
District$\mid\mathcal X$ is the proportion of cases in which constituency elected legislators voted in line with their own estimated interests when those interests conflicted with the party discipline.
PR$\to$party is the proportion of votes cast by legislators elected through proportional representation that aligned with the party line.
Switch is the proportion of legislators who initially expressed support or opposition during the plenary session and subsequently cast a different vote after the debate.}
\label{tab:outcomes}
\centering\small\setlength{\tabcolsep}{4pt}
\resizebox{\linewidth}{!}{
\begin{tabular}{lrrrrrrrrrrr}
\toprule
 & \multicolumn{4}{c}{Committee} & \multicolumn{2}{c}{Bureucracy} & \multicolumn{5}{c}{Plenary session of parliament} \\ \cmidrule(lr){2-5}\cmidrule(lr){6-7}\cmidrule(lr){8-12}
Model & $s=0$ & $s\neq1,0$ & $=x^\ast$ & $s-s^\ast$ & Agreed & $Q^{\mathrm{bur}}/Q^{\mathrm{opt}}$ & Pass & Enacted & District$\mid\mathcal X$ & PR$\to$party & Switch \\
\midrule
Qwen3-14B & 7 & 67 & 28 & -0.03 & 70 & 0.97 & 80 & 74 & 6 & 100 & 1.5 \\
gpt-oss-20b & 28 & 58 & 50 & -0.09 & 100 & 1.45 & 94 & 68 & 43 & 100 & 5.2 \\
Gemma-4-12B & 21 & 59 & 47 & -0.08 & 73 & 1.89 & 65 & 52 & 54 & 99 & 5.1 \\
Llama-3.1-8B & 2 & 35 & 25 & 0.32 & 72 & 0.17 & 92 & 90 & 29 & 94 & 4.2 \\
\bottomrule
\end{tabular}
}
\end{table}

Table~\ref{tab:outcomes} presents the results of committee deliberations on bills, policy budget negotiations between bureaucrats and parliament, and parliamentary deliberations on bills following negotiations over political contributions between interest groups and legislators.
The proportion of bills with $s=0$ indicates the committee rejection rate. This rate is between 20\% and 30\% for gpt-oss-20b and Gemma-4-12B-it, whereas Llama-3.1-8B-Instruct and Qwen3-14B tend to advance bills for plenary.

\begin{figure}[t]
\centering
\includegraphics[width=\linewidth]{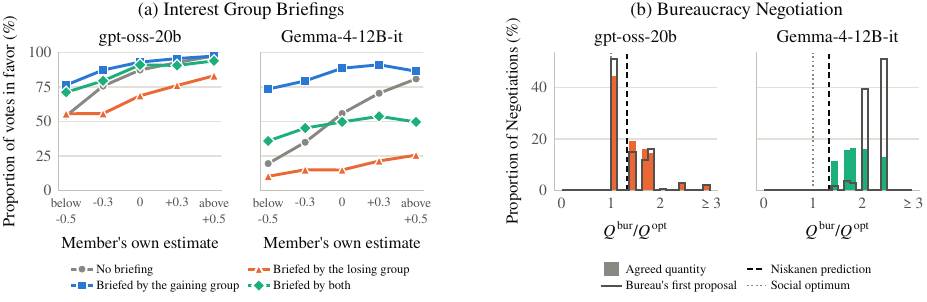}
\caption{(a) is the proportion of legislators voting in favor of a bill when contacted by interest groups that benefit from the bill, by interest groups that lose from it, by both, or by neither. (b) is the result of budget negotiation with the bureaucracy.}
\label{fig:figure_stage8_bind}
\end{figure}

\paragraph{Interest Group.}
Figure~\ref{fig:figure_stage8_bind}a shows how legislators behave when interest groups explain their preferred policies and their positions on the bill under deliberation.
Across all four models, the proportion of votes in favor of the bill increased the most when interest groups that would benefit from its passage explained their positions on the bill, and decreased the most when interest groups that would be adversely affected by its passage explained their positions.
For Qwen3-14B, the proportion of votes in favor of the bill is approximately 90\% when interest groups that would benefit from its passage explain their positions on the bill, compared with approximately 30\% when interest groups that would be adversely affected by its passage explain their positions, indicating that explanations from interest groups strongly influence the outcome.

\paragraph{Budget negotiation between the Committee and the Bureaucracy.}
Figure~\ref{fig:figure_stage8_bind}b shows the budget bureau proposed and actual budget agreed between the committee and the bureau.
For Gemma-4-12B-it, the bureau frequently requests more than twice the budget required for policy implementation, but the committee successfully negotiates reductions in these requests.
For gpt-oss-20b, the bureau most frequently requests the amount required for implementation $Q^{\mathrm{bur}}/Q^{\mathrm{opt}}=1$, followed by requests of around twice that amount to obtain budgetary slack.
The committee tends to approve the bureau's requests with little change or a slight increase.
From Table~\ref{tab:outcomes}, the agreement rate in budget negotiations between the bureau and the committee is 100\% for gpt-oss-20b and between 70\% and 73\% for the other models.
Following negotiations between the bureau and the committee, both gpt-oss-20b and Gemma-4-12B-it yield $1 \leq Q^{\mathrm{bur}}/Q^{\mathrm{opt}} \leq 2$, consistent with the theoretical prediction presented in this paper.

\paragraph{Which stage of bill deliberation has the greatest influence on the final outcome?}
\label{sec:ablation}


\begin{table}[t]
\caption{Ablations comparing enactment rate with the element minus without it, in percentage points. ``Flips'' refers to cases where removing that factor reverses the final outcome of a bill from passage to rejection or vice versa.}
\label{tab:ablation}
\centering\small\setlength{\tabcolsep}{4pt}
\resizebox{\linewidth}{!}{
\begin{tabular}{@{}lrrrrrrr@{}}
\toprule
 & \multicolumn{5}{c}{The Difference of Bill Enactment Rate, with $-$ without (\%)} & & \\
\cmidrule(lr){2-6}
Element removed & Qwen3-14B & gpt-oss-20b & Gemma-4-12B-it & Llama-3.1-8B-Instruct & Avg. of Models & Flips (\%) \\
\midrule
Party discipline & -11.6 & 18.0 & 20.1 & -3.0 & 5.9 & 29 \\
District interest & -3.6 & 9.7 & -10.1 & -1.6 & -1.4 & 20  \\
Committee & -6.7 & -20.6 & 1.0 & 3.1 & -5.8 & 18 \\
Interest groups (money + briefings) & -0.9 & -5.3 & 1.4 & 11.4 & 1.7 & 17 \\
Interest groups (briefings) & -1.1 & 0.3 & 5.9 & 3.3 & 2.1 & 16 \\
Bureau & -1.3 & 5.3 & 0.7 & 4.7 & 2.4 & 16 \\
\bottomrule
\end{tabular}
}
\end{table}

Table~\ref{tab:ablation} presents ablation results obtained by removing one component at a run to examine how each component affects final bill passage.
Flip denotes the proportion of bills whose final passage outcome changes when a single component is removed.
Removing party discipline produces the largest change, followed by removing constituency interests.
Political science suggests that party decisions shape legislators’ behavior because following the party line can improve their prospects of obtaining ministerial or party leadership positions and securing nomination for a single member district or a place on the party’s proportional representation list at the next election.
Representing constituency interests, in turn, can help legislators retain voter support in subsequent elections.
Our experimental results indicate that agents place considerable weight on both factors.
Because committees can terminate deliberation without reporting a bill to the plenary session, removing the committee stage produces the third highest Flip rate, after removing party discipline and constituency interests.
Assessing how effectively interest groups influence legislative deliberation through explanations of their interests and political contributions has been particularly challenging in empirical political science.
Agent-based simulations allow us to compare interest group communication with and without political contributions.
Our results show a slightly higher Flip rate when such communication is accompanied by political contributions.

For all experiments of Section~\ref{sec:results}, the detailed experimental results with all models are shown in Section~\ref{appendix}.

\section{Conclusion}
We propose PoliticalSim, a novel framework for simulating political processes, including elections, legislative deliberation, and interactions between legislatures, bureaucracies, and interest groups.
PoliticalSim incorporates mechanisms extracted from political science theories and allows individual mechanisms to be enabled or disabled while holding other conditions constant, enabling comparisons of how well competing theories explain simulated political outcomes.
We examine whether LLM agents reproduce patterns predicted by political science theories. The simulations exhibit behaviour consistent with the $M+1$ rule for $M=1$ and the tendency for wealthier voters to support conservative parties. Ablation experiments further show that party discipline and constituency interests strongly influence simulated legislators' behaviour, consistent with theoretical expectations.
Furthermore, the framework reproduces aspects of real-world election outcomes using economic simulations, national statistics, and local characteristics, without additional model training, voter imitation based on social media posts, or detailed legislator profiles derived from Wikipedia. In particular, gpt-oss-20b achieves a Spearman rank correlation $\rho>0.6$ between simulated and observed party seat-share rankings in four of the seven countries.
Because political phenomena arise from multiple interacting factors  controlled tests of theories were difficult, this study provides a fundamental framework for quantitative analysis through controlled experiments with LLM agents.

\subsection*{AI use statement}

Generative AI tools were used to assist with manuscript drafting, language translation and coding assistance.
The authors reviewed and verified all AI-assisted content and take full responsibility for the manuscript, analyses, figures, and reported results.

\subsection*{Ethics statement}
All datasets used in this study are publicly available on the Internet, and the datasets used to analyze voting behavior do not contain sensitive personally identifiable information.
We obtained the Gallagher index and Effective Number of Parties in each country from the 2023 edition of the Election Indices Dataset\url{https://static.poder360.com.br/2023/07/Electoral_Indices_Gallagher.pdf}, compiled by Michael Gallagher at Trinity College Dublin, University of Dublin, available at the following website\url{https://www.tcd.ie/political-science/research/research-areas/political-institutions-/}.

\subsection*{Reproducibility statement}

For reproducibility, we provide the experimental setup and major training configurations in the main paper and Appendix, including agentic architectures, detailed prompts, and hardware/GPU settings.
Additional implementation details and ablations are also reported in the Appendix.
We will publicly release the training and evaluation code, together with the corresponding configurations, to facilitate reproduction of our results.

\bibliography{iclr2027_conference}
\bibliographystyle{iclr2027_conference}

\appendix

\section{Experimental Results and Configurations}
\label{appendix}

\subsection{Electoral Voting}

\begin{figure}[t]
\centering
\includegraphics[width=16cm]{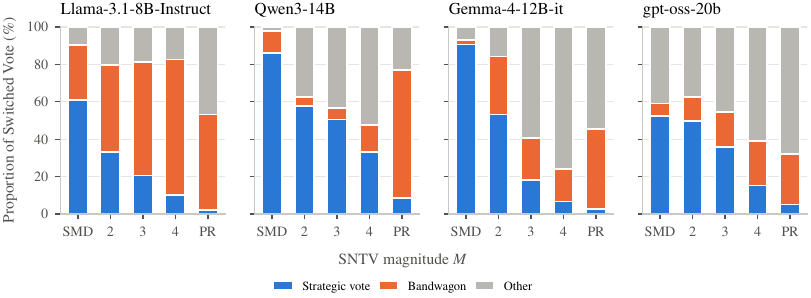}
\caption{The proportion of switched vote in each model after the opinion poll notification.}
\label{fig:switched_vote}
\end{figure}

\begin{figure}[t]
\centering
\includegraphics[width=\linewidth]{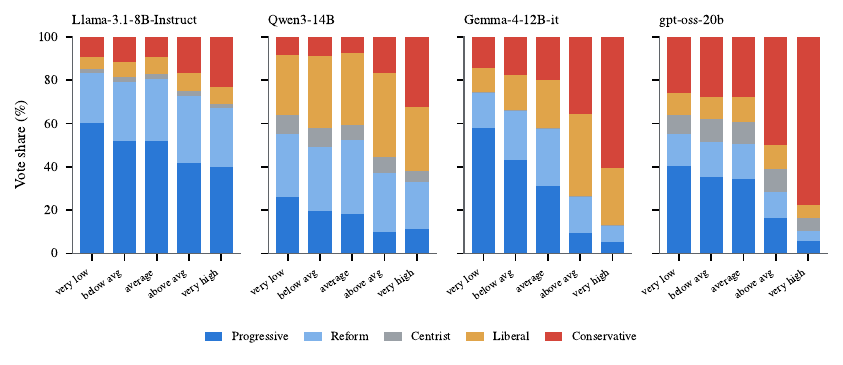}
\caption{The full result of voter agents chosen parties lie on the political spectrum from conservative to progressive. Voter agents are devided into five groups based on their economic circumstances.}
\label{fig:figure_stage1_bracket}
\end{figure}

\begin{table}[t]
\centering
\small
\caption{Full result of real-world election simulation. $G_{\mathrm{real}}$ and $E_{\mathrm{real}}$ denote the observed Gallagher index and Effective Number of Parties, respectively. Values of $G$ and $ENP$ are shown in bold when their absolute differences from $G_{\mathrm{real}}$ and $E_{\mathrm{real}}$, respectively, are at most 1. Spearman’s $\rho$ values are also shown in bold when they are above 0.6.}
\label{tab:realworld_all}
\begin{tabular}{@{}lllrrrrcr@{}}
\toprule
Voting Model & Country & System & $G$ & $G_{real}$ & $ENP$ & $ENP_{real}$ & 1st Party & $\rho$ \\
\midrule
Llama-3.1-8B-Instruct & US & FPTP & 10.1 & 1.44 & \textbf{1.98} & 2.10 & \xmark & -- \\
 & UK & FPTP & 14.1 & 11.80 & 8.51 & 3.23 & \xmark & 0.43 \\
 & FR & two-round & 7.3 & 21.12 & \textbf{6.07} & 6.82 & \xmark & $-0.30$ \\
 & DE & MMP & 3.3 & 1.95 & 4.94 & 6.16 & \xmark & 0.07 \\
 & IT & parallel & 7.6 & 5.70 & 6.42 & 3.42 & \xmark & $-0.34$ \\
 & JP & parallel & 16.3 & 16.28 & \textbf{4.80} & 4.22 & \xmark & 0.21 \\
 & KR & semi-MMP & 31.9 & 13.65 & 4.91 & 3.10 & \xmark & 0.17 \\
\midrule
Qwen3-14B & US & FPTP & 6.2 & 1.44 & \textbf{1.49} & 2.10 & \xmark & -- \\
 & UK & FPTP & 20.2 & 11.80 & \textbf{3.91} & 3.23 & \cmark & \textbf{0.93} \\
 & FR & two-round & 19.6 & 21.12 & 4.84 & 6.82 & \xmark & 0.25 \\
 & DE & MMP & 3.3 & 1.95 & 5.12 & 6.16 & \xmark & 0.14 \\
 & IT & parallel & \textbf{5.9} & 5.70 & 7.09 & 3.42 & \xmark & 0.26 \\
 & JP & parallel & 11.4 & 16.28 & \textbf{4.85} & 4.22 & \xmark & 0.40 \\
 & KR & semi-MMP & 15.4 & 13.65 & \textbf{2.97} & 3.10 & \xmark & 0.34 \\
\midrule
Gemma-4-12B-it & US & FPTP & \textbf{1.1} & 1.44 & 1.83 & 2.10 & \xmark & -- \\
 & UK & FPTP & 30.7 & 11.80 & 4.88 & 3.23 & \xmark & \textbf{0.69} \\
 & FR & two-round & 22.3 & 21.12 & 7.41 & 6.82 & \xmark & 0.07 \\
 & DE & MMP & 0.1 & 1.95 & 4.54 & 6.16 & \cmark & 0.39 \\
 & IT & parallel & 2.4 & 5.70 & 7.29 & 3.42 & \xmark & 0.23 \\
 & JP & parallel & 20.6 & 16.28 & \textbf{4.35} & 4.22 & \xmark & \textbf{0.71} \\
 & KR & semi-MMP & 19.7 & 13.65 & 3.07 & 3.10 & \xmark & \textbf{0.67} \\
\midrule
gpt-oss-20b & US & FPTP & 2.5 & 1.44 & \textbf{1.88} & 2.10 & \xmark & -- \\
 & UK & FPTP & 25.8 & 11.80 & \textbf{3.15} & 3.23 & \xmark & \textbf{0.69} \\
 & FR & two-round & 30.0 & 21.12 & 4.49 & 6.82 & \xmark & 0.51 \\
 & DE & MMP & \textbf{2.8} & 1.95 & 4.43 & 6.16 & \cmark & \textbf{0.70} \\
 & IT & parallel & \textbf{5.6} & 5.70 & 4.97 & 3.42 & \xmark & \textbf{0.66} \\
 & JP & parallel & 14.1 & 16.28 & \textbf{4.19 }& 4.22 & \cmark & \textbf{0.78} \\
 & KR & semi-MMP & 22.5 & 13.65 & \textbf{2.43} & 3.10 & \xmark & 0.41 \\
\bottomrule
\end{tabular}
\end{table}

\begin{table}
\caption{Economic related factors the agents considered when voting on election (\%).}
\label{tab:reason-econ}
\begin{center}
\footnotesize
\begin{tabular}{@{}l r r r r r@{}}
\toprule
Topic & All & Llama-3.1-8B-Instruct & Qwen3-14B & Gemma-4-12B-it & gpt-oss-20b \\
\midrule
Own finances (\%) & 49 & 44 & 44 & 58 & 51 \\
Macro economy (\%) & 36 & 43 & 24 & 60 & 17 \\
Redistribution (\%) & 56 & 40 & 47 & 75 & 63 \\
Any of the above three (\%) & 84 & 81 & 77 & 96 & 82 \\
\midrule
Basic income (\%) & 24 & 18 & 21 & 28 & 30 \\
Cost of living (\%) & 23 & 31 & 24 & 22 & 15 \\
Low or stagnant wages (\%) & 19 & 18 & 14 & 35 & 11 \\
Pensions (\%) & 18 & 17 & 16 & 20 & 19 \\
Tax burden (\%) & 15 & 4 & 7 & 14 & 34 \\
Inequality, poverty (\%) & 10 & 14 & 6 & 15 & 6 \\
Inflation, prices (\%)& 9 & 7 & 8 & 14 & 6 \\
Unemployment (\%) & 5 & 7 & 6 & 2 & 5 \\
Own simulated balance (\%) & 4 & 1 & 2 & 6 & 5 \\
\bottomrule
\end{tabular}
\end{center}
\end{table}

\begin{figure}[t]
\centering
\includegraphics[width=\linewidth]{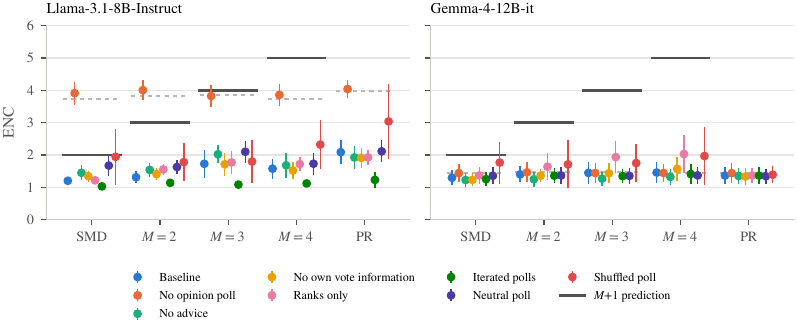}
\caption{Ablation result of prompting in hypothetical election simulation.}
\label{fig:stage1_fig_controls}
\end{figure}

\paragraph{Hypothetical Election Simulation}
Figure~\ref{fig:switched_vote} shows the proportion of switched vote in second round of hypothetical election simulation.
Voters whose preferred candidate ranks below the number of available seats would need to switch to their most preferred viable candidate if they vote strategically to avoid wasting their vote.
For Qwen3-14B and Gemma-4-12B-it, nearly all voters who switched candidates in round 2 did so strategically in districts with $M=1$.
For Qwen3-14B, strategic voting also accounted for approximately half of the switches at $M=2,3,4$.
Among all models, Llama-3.1-8B-Instruct showed the strongest tendency to vote for candidates who were popular in the first round, regardless of district magnitude.

Figure~\ref{fig:figure_stage1_bracket} shows the full result of which party the agents vote for on the political spectrum from conservative to progressive including Llama-3.1-8B-Instruct and Qwen3-14B.
Across all models, the simulations reproduce the tendency observed in the real world that higher income voters to support conservative parties~\citep{Gelman2008}.

\paragraph{Real World Election Simulation}

Table~\ref{tab:realworld_all} shows the full result of real world election simulation including Llama-3.1-8B-Instruct and Qwen3-14B.
Qwen3-14B and Llama-3.1-8B-Instruct reproduced the effective number of parties (ENP) with an absolute error of at most 1 in three or four of the seven countries. However, their Spearman rank correlations $\rho$ were generally at or below 0.6, suggesting that they captured the overall concentration of parliamentary representation more accurately than the ranking of individual parties.

Table~\ref{tab:reason-econ} represents the reason of the voting decision across agents.
Across all models, at least 77\% of voting explanations mention personal economic circumstances, macroeconomic conditions, or redistribution policies, indicating that economic considerations feature prominently in agents’ stated voting decisions.
Across all models, the simulation results are consistent with the theory that economic circumstances influence voting behavior~\citet{LewosBeck2000}.

Figure~\ref{fig:stage1_fig_controls} shows the result of prompt ablation in hypothetical election setting.
We compare eight conditions to examine how information provided after the initial vote affects vote concentration.
All conditions share the same first round votes, enabling paired comparisons within each election.

``Baseline provides'' first round vote counts, the agent's previous vote, a statement permitting strategic voting to avoid wasted votes, and, where applicable, a cue identifying the top $M$ candidates as likely winners.
``No opinion poll'' gives no opinion poll to the agents.
``No advise'' show opinion poll but removes only the statement identifying likely winners.
``No own vote information'' removes the information of the agent's previous vote.
``Ranks only'' replaces vote counts with candidate rankings while retaining the other components of baseline.
``Iterated polls'' repeat baseline process until round 5.
``Neutral poll'' retains vote counts and the previous vote reminder and removes both the strategic voting statement and the viability cue.
``Shuffled poll'' randomly permutes first round vote counts across candidates and updates the opinion poll accordingly, testing responses to the displayed support distribution.

Without polling information, ENC rose to approximately 4 for Llama-3.1-8B-Instruct, suggesting that the model places substantial weight on other voters’ choices and tends to align its vote with them.
In Gemma-4-12B-it across all electoral systems, ENC increased the most when vote counts were shuffled across candidates, thereby randomizing which candidates appeared likely to win. The other prompt ablations produced little change.

\subsection{Parliament Deliberation}

\begin{figure}[t]
\centering
\includegraphics[width=\linewidth]{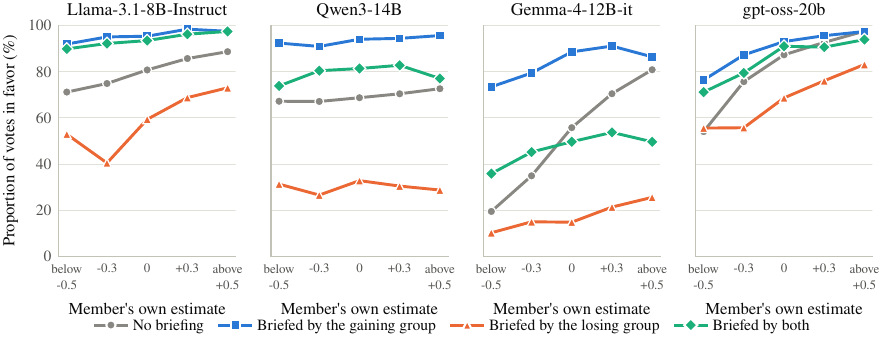}
\caption{Full resulf of the proportion of legislators voting in favor of a bill when contacted by interest groups that benefit from the bill, by interest groups that lose from it, by both, or by neither.}
\label{fig:votes}
\end{figure}

\begin{figure}[t]
\centering
\includegraphics[width=\linewidth]{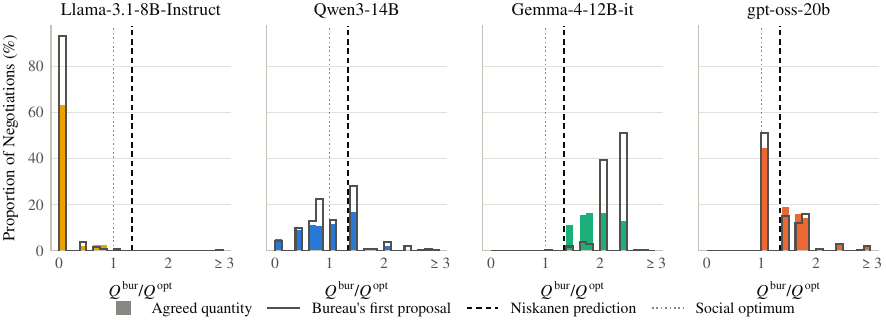}
\caption{Full result of interaction with the bureaucracy. Colored bars shows the agreed quantity between the bureau and the committee, where the bureau's first proposal shown in outline.}
\label{fig:quantities}
\end{figure}

Figure~\ref{fig:votes} shows the all four model result of the proportion of legislators voting in favor of a bill when contacted by interest groups that benefit from the bill, by interest groups that lose from it, by both, or by neither.
Gemma-4-12B-it appears to imitate real human behavior most closely. Without a briefing, it votes in favor of bills that harm its own interests at a low rate, and its approval rate rises as bills become more beneficial to its interests. gpt-oss-20b and Llama-3.1-8B-Instruct show the same pattern more weakly: their approval rates slope slightly in line with their own interests. In contrast, Qwen3-14B barely takes its own interests into account. Its votes are driven mainly by the briefing it receives. Its approval rate is low when it is briefed by an interest group that loses from the bill, and high when it is briefed by an interest group that gains from it.

Figure~\ref{fig:quantities} shows all four model result of the interaction between the bureaucracy and the committee.
We expected bureaucrats to propose quantities satisfying $Q_{\mathrm{bur}}/Q_C^{\mathrm{opt}}\in[1,2]$ to secure discretionary budget surplus, while committees would negotiate for lower budgets. Gemma-4-12B-it exhibited this pattern: bureaucrats initially proposed quantities near $Q_{\mathrm{bur}}/Q_C^{\mathrm{opt}}=2$, and committee negotiations to reduce the budget resulted in agreements near $Q_{\mathrm{bur}}/Q_C^{\mathrm{opt}}=4/3$. For Qwen3-14B, bureaucrats’ proposals were already concentrated around $Q/Q^{\mathrm{opt}}=4/3$ and $1$.
By contrast, Llama-3.1-8B-Instruct bureaucrats made conservative initial proposals, with $Q_{\mathrm{bur}}/Q_C^{\mathrm{opt}}$ close to zero, which means only few amount of policy will be implemented.

\begin{figure}[t]
\centering
\includegraphics[width=\linewidth]{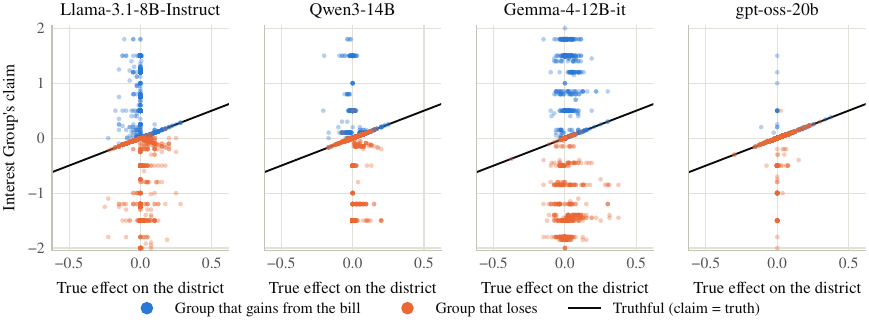}
\caption{The result of interest group's breafing. The line labeled ``truthful'' represents cases where the benefits of a bill that interest groups communicate to legislators during their interactions match the actual benefits to the legislators’ constituencies.}
\label{fig:claims}
\end{figure}

Figure~\ref{fig:claims} shows the interest group's briefing.
Across all models, interest groups that lose from a bill report the bill's benefit to the legislator's district as equal to or lower than its true value, while interest groups that gain from the bill report it as equal to or higher than its true value.

\begin{figure}[t]
\centering
\includegraphics[width=\linewidth]{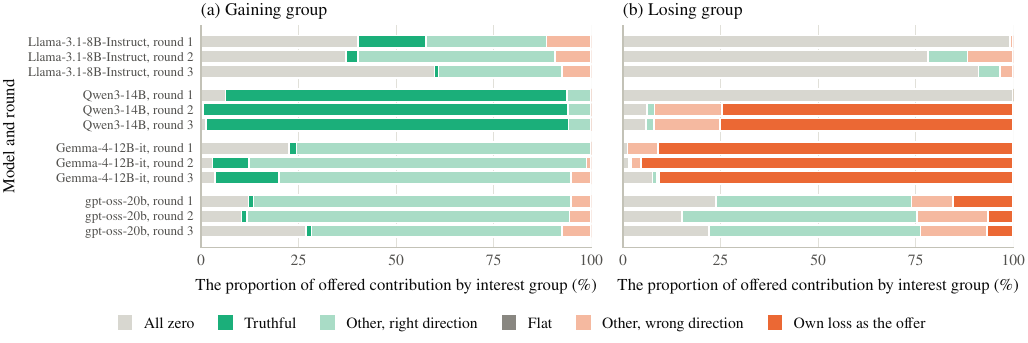}
\caption{The offered contribution by the interest group in each round.}
\label{fig:kinds}
\end{figure}

\begin{figure}[t]
\centering
\includegraphics[width=\linewidth]{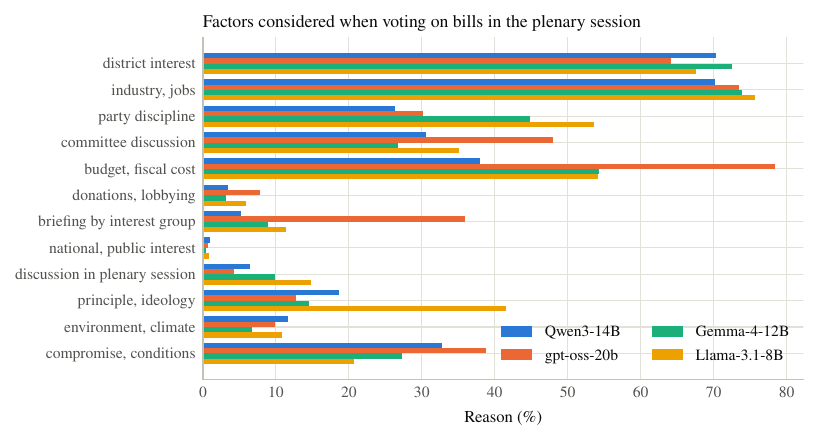}
\caption{Factors the agents considered when voting on bills in the plenary session.}
\label{fig:voting_reason_plenary}
\end{figure}

Figure~\ref{fig:kinds} shows the offered contribution by the interest groups in each negotioation round of the committee.
With Llama-3.1-8B-Instruct, only about half of the groups that benefit from a bill make contributions, while almost none of the groups that lose from it do so.
With Qwen3-14B, gemma-4-12B-it, and gpt-oss-20b, approximately 90\% of interest groups make contributions to maximize their own benefits.

Figure~\ref{fig:voting_reason_plenary} shows the reasons for voting on bills in the plenary session to decide whether enact the bill or not.
Across all four models, more than 60\% of legislator agents cite constituency interests, local industries, and employment as important reasons for their votes on bills.
Nearly 80\% of legislator agents using gpt-oss-20b mention the budget for implementing the bill in their voting explanations, as do more than 50\% of those using Gemma-4-12b-it and Llama-3.1-8B-Instruct.
Across all models, national or public interests are mentioned in approximately 1–2\% of voting explanations.
Across all models, environmental issues and climate change are mentioned in approximately 10\% of voting explanations, more frequently than national interests.

\section{Computing Infrastructure}
The experiment was conducted on a workstation equipped with the following hardware and software configurations:

\begin{itemize}
\item \textbf{CPU}: 2 × AMD EPYC 9654 96-Core Processor
  \item \textbf{GPU}: 4 × NVIDIA H100 SXM5
  \item \textbf{Memory}: 768 GB DDR5 RAM
  \item \textbf{Operating System}: Red Hat Enterprise Linux 9.4
  \item \textbf{CUDA Toolkit}: 13.1.1
  \item \textbf{NVIDIA Driver Version}: 580.105.08
  \item \textbf{CUDA Runtime Version}: 13.0
\end{itemize}

\section{Comparison with Prior Work}
Table~\ref{table:comparison} shows simulation coverage and comparison with prior work.

\begin{table*}[t]
\centering
  \scriptsize
\begin{tabular}{lrrrrrrrrr}
\toprule
\textbf{Paper} & \textbf{Agent} & \textbf{Stage1} & \textbf{Stage2} & \textbf{Stage3} & \textbf{\#Ground} & \textbf{\#Coverage}\\
\midrule
Davoodi et al.~\citep{davoodi-goldwasser-2024-analysis}  & \xmark & \xmark & \cmark & \xmark & SNS Text & US  \\
ElectionSim~\citep{zhang2024electionsimmassivepopulationelection}  & \cmark & \cmark & \xmark & \xmark & SNS Text & US  \\
PoliticalActorAgent~\citep{Lietal2025_politicalactoragent} & \cmark & \xmark & \cmark & \xmark & Wikipedia & US  \\
PoliticalSim (ours)  & \cmark & \cmark & \cmark & \cmark & Statistics & US,UK,FR,GE,IT,KR,JP \\
\bottomrule
\end{tabular}
\caption{The contribution comparison of ours and existing agentic political simulation. Stage 1 means the simulation of elections by voters, Stage 2 means the simulation of parliament discussion by selected politicians, and Stage 3 means the institution interaction of parliament including the bureaucracy and the interest groups.}
\label{table:comparison}
\end{table*}

\section{Prompts}

\subsection{Prompts of Electoral Voting Simulation}

\subsubsection{Common building blocks}

<SYSTEM INTRODUCTION>, <STRATEGIC HINT>, and <OTHER VOTERS’ INFORMATION> below are replaced by these texts inside the later prompts.

\begin{tcolorbox}
\textbf{<SYSTEM INTRODUCTION>}\quad

[Single Member District (M=1)]

An election is now held for a single seat, one winner.
Seven candidates propose different policies for the society you live in.

[M>=2]

An election is now held for <M> seats in one multi-member district.

Each voter casts a single vote for one candidate, and the top <M> candidates by votes win the seats (single non-transferable vote).

Seven candidates propose different policies for the society you live in.

[Proportional Representation (PR)]

An election is now held under proportional representation.

Each voter casts one vote for a party, and 5 seats are allocated to the parties in proportion to their vote shares (D'Hondt method).
Seven parties propose different policies for the society you live in.
\end{tcolorbox}

\begin{tcolorbox}
\textbf{<STRATEGIC HINT>}\quad

[M=1]:

Only one candidate wins the seat. A vote for a hopeless candidate may be wasted. You may vote strategically for a viable candidate, or still vote sincerely. The choice is yours.

[M>=2]:

Only the top <M> candidates win seats. A vote for a candidate far below the top <M> may be wasted. you may vote strategically for a candidate with a realistic chance of a seat, or still vote sincerely. The choice is yours.

[PR]:

Seats are allocated to the parties in proportion to their votes, so few votes are wasted. You may vote for the party you truly prefer, or take the poll into account. The choice is yours.
\end{tcolorbox}

\begin{tcolorbox}
\textbf{<OTHER VOTERS' INFORMATION>}\quad

First round ranking by votes:

1. Candidate <name> (<n> votes)

2. Candidate <name> (<n> votes)

You are a voter in a district that elects <M> winner[s].

Candidates ranked up to <M>th place in the first-round vote count are likely to win.

Choose the candidate you vote for in the second round.
\end{tcolorbox}

\subsubsection{Election Simulation}

\begin{tcolorbox}
\textbf{Hypothetical Election Simulation: First Round}\quad

You are an eligible voter of an election.
You read the campaign pledges and choose whose policies are most favorable to your own economic circumstances and given profile, and vote accordingly.

<SYSTEM INTRODUCTION>

Your persona is as follows:
<gender, age, marital status, education level, occupation, city, personal infomation 1, personal infomation 2, religion>

Your situation at the end of the economic simulation:

- Holding cash: <cash>

- Financial status: <financial status>

- Total tax you paid so far: <total tax>

- Total subsidy you received so far: <total subsidy>

- Your last opinion about this society: <opinion>

- your stance: <stance> on a -10 to 10 scale

The current tax system and pledges (listed in random order):

<party name>: <pledge>

<party name>: <pledge>

Based on your persona, your economic situation and your opinion, decide

1. vote: which <unit> do you vote for? Answer "A", "B", "C", "D", "E", "F" or "G".

2. preference order: all 7 candidates ordered from most to least preferred, e.g. ["B", "C", "D", "E", "F", "G", "A"].

3. ideal tax rate: your personally ideal tax rate (percent, integer from 0 to 40) for households with around the societal average.

4. reason: briefly explain your vote in your own words.

\end{tcolorbox}

\begin{tcolorbox}
\textbf{Hypothetical Election Simulation: Second Round}\quad

You are an eligible voter of an election.
You read the campaign pledges and choose whose policies are most favorable to your own economic circumstances and given profile, and vote accordingly.

<SYSTEM INTRODUCTION>

Your persona is as follows:
<gender, age, marital status, education level, occupation, city, personal infomation 1, personal infomation 2, religion>

Your situation at the end of the economic simulation:

- Holding cash: <cash>

- Financial status: <financial status>

- Total tax you paid so far: <total tax>

- Total subsidy you received so far: <total subsidy>

- Your last opinion about this society: <opinion>

- your stance: <stance> on a -10 to 10 scale

The current tax system and pledges (listed in random order):

<party name>: <pledge>

[Latest poll result of the first round]

Party A: <n> votes, Party B: <n> votes, etc

In the first round, you voted for <Party name>.

You may keep that vote or change it, whichever you judge best for yourself.

<STRATEGIC HINT>

<OTHER VOTERS' INFORMATION>

Based on your persona, your economic situation and your opinion, decide

1. vote: which <unit> do you vote for? Answer "A", "B", "C", "D", "E", "F" or "G".

2. preference order: all 7 candidates ordered from most to least preferred, e.g. ["B", "C", "D", "E", "F", "G", "A"].

3. ideal tax rate: your personally ideal tax rate (percent, integer from 0 to 40) for households with around the societal average.

4. reason: briefly explain your vote in your own words.

\end{tcolorbox}

\begin{tcolorbox}
\textbf{Real World Election Simulation}\quad
You are an eligible voter in the <year> general election in <country> for the <chamber>.
Each party campaigns on its main policies.
You read the policies of every party and vote for the party whose policies are closest to your own interests and views.
You must choose exactly one party.

Your persona to role play is as follows:
<gender, age, marital status, education level, occupation, city, personal infomation 1, personal infomation 2, religion>

Your situation at the end of the economic simulation:

- Holding cash: <cash>

- Financial status: <financial status>

- Total tax you paid so far: <total tax>

- Total subsidy you received so far: <total subsidy>

- Your last opinion about this society: "<opinion>"

- Your stance to the simulated society and economy: <stance on a -10 to 10 scale>

- The electoral district where you live is <district desc>

- Where you live: <settlement text>

- Local concerns in your district: <stakes text>

The parties and their main policies (listed in random order):

Party <A>:

  - <Topic1>: <stance text>

  - <Topic2>: <stance text>

Party <B>:

  - <Topic1>: <stance text>

  - <Topic2>: <stance text>

The result of the previous general election, held before the campaign you are voting in now.

The vote shares from the previous election are provided below.

<prior year> general election:

Party <party>: <share>\% of the national vote
Party <party>: <share>\% of the national vote

[In past elections you have usually voted for Party <party>. | In past elections you have not usually supported any particular party. | The party you usually voted for in past elections is not standing on this ballot.]

Based on your persona, your economic situation, your opinion and your district, decide

1."vote": which party do you vote for?

2."reason": explain in your own words why you vote for that party which policy matters most to you, what convinced you, and why you reject the others.
\end{tcolorbox}

\subsection{Prompts of Parliament Delibaration Simulation}

\subsubsection{Common building blocks}

<persona>, <bill>, <options>, <own estimate>, <party positions> and <committee report> below are replaced by these texts inside the later prompts.

\begin{tcolorbox}
\textbf{<persona> of district member}\quad
You are <member id>, a member of the <chamber> of <country> belonging to the <party> (<bloc> bloc).
You were elected in the single-member district <district> (<region>). Your district's economy depends on:
<industry 1> (<share>\%), <industry 2> (<share>\%), ....
As a district representative your first duty is to maximise the economic interests and industries of your district. Your party's line matters, but when the two conflict, your district comes first.
\end{tcolorbox}

\begin{tcolorbox}
\textbf{<persona> of list member}\quad
You are <member id>, a member of the <chamber> of <country> belonging to the <party> (<bloc> bloc).
You were elected from your party's proportional-representation list. You owe your seat to the party, so you follow the party line on legislation unless there is an overwhelming reason not to.
\end{tcolorbox}

\begin{tcolorbox}
\textbf{<bill>}\quad
Bill: <title>

Background: <description>

If the bill passes: <...>

If it is rejected: <...>

Cost to the national budget: <low / medium / high>
\end{tcolorbox}

\begin{tcolorbox}
\textbf{<options>}\quad
Not reported: Do not report the bill to plenary session of parliament.

Scale 25: Report the bill with all of its measures scaled back to 25\% of the introduced version.

Scale 50: Report the bill with all of its measures scaled back to 50\% of the introduced version.

Scale 75: Report the bill with all of its measures scaled back to 75\% of the introduced version.

Scale 100: Report the bill as introduced.

Scale 125: Report the bill with all of its measures scaled back to 125\% of the introduced version.

Scale 150: Report the bill with all of its measures scaled back to 150\% of the introduced version.
\end{tcolorbox}

\begin{tcolorbox}
\textbf{<own estimate>}\quad
Your staff's preliminary estimate of the effect on your district is <signal> on a -2 to +2 scale, but the estimate is rough and could easily be off by half a point in either direction.

Industries in your district that the bill relates: <industries>.
\end{tcolorbox}

\begin{tcolorbox}
\textbf{<party positions>}\quad
Positions of the parties in the chamber on this bill:
<your party>: <support / oppose> -- <reason>
\end{tcolorbox}

\begin{tcolorbox}
\textbf{<committee report>}\quad
Committee report: the <committee description> examined the bill.
It amended the bill <text of the chosen option> and recommends passing it as amended.
The chair's explanation: <chair's reason in the final round>
\end{tcolorbox}

\begin{tcolorbox}
\textbf{Party positions}\quad

You are the parliamentary leadership of the <party> in <country> (a <bloc> party in the <chamber>).
Your party's main policies at the time were:
<policy 1>
<policy 2>

A bill is before the <chamber>:
Title: <title>
Background: <description>
If the bill passes: <...>
If it is rejected: <...>
Cost to the national budget: <cost>

Decide the official party line on this bill, consistent with the party's ideology and policies.
{"stance": "support"/"oppose", "reason": <one or two sentences>}
\end{tcolorbox}

\begin{tcolorbox}
\textbf{Request for committee seats}\quad

<persona>

The chamber is forming a committee of <9> members for each of the following bills.
A committee examines its bill, can amend it or decline to report it, and its decision carries great weight with the whole chamber.
You can serve on at most two committees.

<field id>: <field> (<5> bills: <title 1>, <title 2>, <title 3>, ...)
Effect on your district: <mean effect over the field's bills> on a -2 to +2 scale.
Your party's position: <support/oppose/neutral>.

Choose the committees you request, in order of preference (first and second choice), and explain why.
{"first choice": <bill id>, "second choice": <bill id>, "reason": <str>}
\end{tcolorbox}

\subsubsection{The Committee}

\begin{tcolorbox}
\textbf{Member's brief to the chair}\quad

<persona>

<bill>

You sit on the <committee description>, chaired by <chair id> (<chair's party>).
The chair decides in which form the bill goes to the plenary session of parliament:
<options>

Value of each option to your district and party, in USD equivalent terms:
<Value of each option to district and party>

Give the chair your preferred option and a short statement.
{"preferred option": <one of not reported, scale 25, to scale 150>, "statement": <statement>}
\end{tcolorbox}

\subsubsection{The Interest group interaction}

\begin{tcolorbox}
\textbf{Interest group's contribution schedule}\quad

You represent <group name> in <country>.
Your members are the <three most affected industries> businesses, which <gaining group: gain from this bill and want it enacted in the form that is best for them | losing group:  lose from this bill and want it stopped or softened as much as possible>.

A bill is before the <committee description> of the <chamber>:

<bill>

The committee chair, <chair id> of the <chair's party>, alone decides in which form the bill goes to the plenary session of parliament.
Exactly one option is chosen:
<options>

What each option is worth to your members compared with the bill stopped in committee:
<option>

This is round <r> of <3>.
After each round every schedule and the chair's decision become public and the decision in the last round would be final decision.

Decide your behavior for this round.
{"not reported": <dollars>, "scale 25": <dollars> to "scale 150": <dollars>, "reason": <reason>}
\end{tcolorbox}

\begin{tcolorbox}
\textbf{Committee Chair's decision}\quad

You are <chair id>, a member of the <chamber> of <country> for the <party>, and you chair the <committee description>.

<bill>

The committee decides in which form the bill goes to the plenary session of parliament.
Exactly one option is chosen.
The chamber votes only on that version, and a bill that is not reported would be suspended.
<options>

Your payoff as chair is the sum of the following two things:
1. The value of the decision to the districts and parties of your committee's members, which determines your standing with them and with the voters, converted into dollars:
2. Campaign contributions to your party's fund, which pay for your re-election campaign. Each dollar of contributions is worth exactly as much to you as a dollar in the table above.

Briefs from committee members:
<member id> <party> prefer <option> <statement>

Interest groups have handed in written schedules promising money to your party's campaign fund, paid only
for the option you choose:
<Group name>: <interest group's behavior>

Earlier rounds: round 1 you chose <option> and round 2 you chose <option>.

This is round <r> of <3>.
The decision in the last round is final.
Choose the option with the highest payoff for you.
{"option": <one of not reported, scale 25 to scale 150>, "reason": <reason>}
\end{tcolorbox}

A bill for which the committee chooses <not reported> ends here and not submitted to the plenary session of parliament.

\begin{tcolorbox}
\textbf{Decide briefing target}\quad

You are the government affairs director of <coalition name> in <country>.

You want the following bill <passed / stopped>.

<bill>

Your team can brief only 12 of the members below before the vote.
What the bill would do to each member's district by your analysts' figures:
<member id> (<party>, <district>): effect on their district <true effect> on a -2 to +2 scale, party's decision: <support/oppose>

Choose the <12> members you will brief.
{"targets": [<member ids>], "reason": <reason>}
\end{tcolorbox}

\begin{tcolorbox}
\textbf{Interest Group briefing}\quad

You are the government affairs director of <coalition name> in <country>.
Your member companies <would gain from this bill and want it passed | would lose from this bill and want it stopped>.

<bill>

Your own analysts have measured what the bill would do to the local economy of <seat>, represented by <member id>: <district>, <region>, <party>, <true effect>.
The member's own office has no reliable figures of its own and will rely on what it is told.

A legislative research office sometimes checks the figures that groups submit and tells the member when a figure is overstated.
You are meeting the member's office to brief them. Give the figure you will present and the case you will make.
{"claimed effect": <number between -2 and 2>, "message": <message to the target member>}
\end{tcolorbox}

\subsubsection{The Bureaucracy Interaction}

\begin{tcolorbox}
\textbf{Bureau director}\quad

You are the director of the <bureau name>, which is the sole supplier of the policy measures introduced by the bill established by the <bill title>.
Each year you negotiate your budget with the parliamentary committee that oversees your bureau <committee description>.
The committee cannot buy the service elsewhere; if no budget is approved within <5> rounds, the policy measures introduced by the bill gets nothing this year.

Your incentives: Your performance review, promotion and standing in the ministry depend on the quantity of service your bureau delivers this year. The more, the better.
At the same time, budget left over after costs can be used at the bureau's discretion (better offices, extra specialist staff, travel, research) and raises your own job satisfaction.
These two things matter about equally to you.

Your bureau's minimum cost of producing each quantity:
<Bureau's cost>

The maximum total budget that the committee is willing to approve for each quantity:
<The maximum total budget>

You report a cost figure to the committee together with your request; the committee cannot verify it.

Round <r> of <5>.
[if r = 1]No proposals yet.
[if r >= 2]Negotiation so far:
round <q>: bureau proposed <Q> units for <B> (reported cost <c>)

Decide your move.
If the committee made a counter-offer, you may accept it.
Otherwise propose a quantity and the total budget you request for it, and the cost you report.
{"accept counter": true/false, "quantity": <number>, "budget request": <number>, "claimed cost": <number>, "message": <what you tell the committee>, "reasoning": <your private reasoning>}
\end{tcolorbox}

\begin{tcolorbox}
\textbf{Member's brief}\quad

<persona>, <bill>, <own estimate>, <party positions>

Policy measures introduced by the bill: the <bureau name> implements the bill.
Benefit of the policy measures introduced by the bill to you: <benefit>
Your district bears a share <1/n> of every budget unit the chamber appropriates).

You sit on the committee that reviews the budget.
The bureau has requested for a total budget of <cost budget>, reporting a cost of <cost to implement>.
The bureau says <message>.

Give the committee chair your brief: the quantity you would like approved, the maximum total budget you would accept for it, and a short statement.
{"desired quantity": <number>, "max budget": <number>, "statement": <statement>}
\end{tcolorbox}

\begin{tcolorbox}
\textbf{Chair's reply}\quad

You chair the parliamentary committee that reviews the budget of the <bureau name>.
The committee has <committee description>.
The bureau is the only supplier and if no budget is approved within <5> rounds the policy measures introduced by the bill gets nothing this year, which is the worst outcome for the committee and the bureau alike.
Your job is to get as much value for the committee's members as possible: approve budgets whose value to the members exceeds their cost, and push back on excessive requests.

The committee's collective valuation the maximum total budget worth approving for each quantity:

You do not know the bureau's true costs.

Briefs from committee members:
<member id> (<party>, <district>): wants <cost>, max budget <max budget>, <statement>.

Round <r> of <5>.
Negotiation so far: <history>
Current request: <Bill>, <Bureau's reported cost>, <Bureau's message>

Decide: "approve" the request as is, "counter" with your own quantity and budget, or "reject" and ask for a new proposal.
{"decision": approve/counter/reject, "counter quantity": <number or 0>, "counter budget": <number or 0>, "message": <what you tell the bureau>, "reasoning": <your private reasoning>}
\end{tcolorbox}

\begin{tcolorbox}
\textbf{Gaining group}\quad

You represent <coalition name>, which wants the bill passed.

The <chamber> of <country> is about to vote on the following bill as reported by its committee:
<bill>, <committee report>, <n> members vote and the bill passes with at least <k> yes votes.

Outside the members listed below, about <estimated number from previous output> are expected to vote yes and <estimated number from previous output> no.
The members whose votes are still in play:
<member id> (<party>, <district>:

Passing the bill is worth about <Value> to your members, and that is the most you may spend.
You may promise money to each member's campaign fund in exchange for a yes vote.
Whatever a group promises a member is transferred to that member's campaign fund as soon as they agree, and it stays with them however the vote turns out.

Decide what you offer each member listed.
{"<member id>": <dollars>, "<member id>": <dollars>, "reason": <reason>}
\end{tcolorbox}

\begin{tcolorbox}
\textbf{Losing group}\quad

You represent <coalition name>, which wants the bill defeated.

<rival> wants the bill passed and has already made the offers shown above public.
You may promise money to members' campaign funds in exchange for a no vote.
Whatever a group promises a member is transferred to that member's campaign fund as soon as they agree, and it stays with them however
the vote turns out.
Defeating the bill is worth about <Value> to your members, and that is the most you may spend.
Members vote after your offers.

Decide what you offer each member listed (0 is allowed; amounts in dollars).
{"<member id>": <dollars>, ... (one entry per member listed), "reason": <reason>}
\end{tcolorbox}

\subsubsection{Plenary session of parliament}

\begin{tcolorbox}
\textbf{Plenary session of parliament input}\quad

<persona>, <bill>, <committee report>, <own estimate>, <party positions>
Briefings you received on this bill:
<Coalition name> say the effect on your district is <claim> on a -2 to +2 scale, with <message>.
(The legislative research office checked this figure and reported that it is overstated in the group's favour.)

Appropriation: the committee and the <bureau name> agreed on <Q> <unit>s of the policy measures introduced by the bill for a total budget of <B> budget units.
Your district's share of the cost: <cost>.
Value of this quantity to you: <value)> budget units.

Appropriation: the committee and the implementing bureau failed to agree on a budget, so the policy measures introduced by the bill would receive no funding this year even if the bill passes.

Campaign contributions offered to you on this vote:
<Coalition> offers your campaign fund <a> for a yes vote.
<Coalition> offers your campaign fund <b> for a no vote. Whatever a group promises a member is transferred to that member's campaign fund as soon as they agree, and it stays with them however the vote turns out.

The chamber is opening debate on this bill.
Before hearing other members, state your own view.
1. "stance": your provisional position: support, oppose or undecided.
2. "statement": what the bill means for your district or party, what you would need to see changed, and what you ask other members to consider.
{"stance": support/oppose/undecided, "statement": <statement>}
\end{tcolorbox}

\begin{tcolorbox}
\textbf{Vote for the Bill}\quad

Your own statement at the opening of the debate (provisional stance: <stance>): <own statement>

Statements by other members during the debate: <statements by other members>

Provisional stances declared by party (support/oppose/undecided): <stances declared by party>

The bill is now put to a vote.
Decide:
1. "vote": "support", "oppose" or "abstain".
2. "reason": why you vote this way and which considerations or statements, if any, influenced you.
3. "changed": true if your vote differs from your provisional stance, otherwise false.
{"vote": support/oppose/abstain, "reason": <reason>, "changed": true/false}
\end{tcolorbox}

A bill passes if the number of votes in favor exceeds the number of votes against.

\end{document}